\documentclass[%
 reprint,
 twocolumn,
superscriptaddress,
 amsmath,amssymb,
 aps,
]{revtex4-2}

\usepackage{graphicx}
\usepackage{dcolumn}
\usepackage{bm}
\allowdisplaybreaks
\usepackage[breaklinks]{hyperref}
\hypersetup{colorlinks=true, linkcolor=blue, citecolor=blue, filecolor=blue, urlcolor=blue}
\graphicspath{{figures}}
\usepackage{mathtools}

\newcommand{\ket}[1]{| {#1} \rangle}
\newcommand{\bra}[1]{\langle {#1} |}
\newcommand{\Tr}{\mathrm{Tr}}

\usepackage[breaklinks]{hyperref}
\hypersetup{colorlinks=true, linkcolor=blue, citecolor=blue, filecolor=blue, urlcolor=blue}
\usepackage[normalem]{ulem}

\begin{document}

\title{Stack operation of tensor networks }
\author{Tianning Zhang}
\email{tianning{\_}zhang@mymail.sutd.edu.sg}
\affiliation{
    Science, Mathematics and Technology, Singapore University of Technology and Design, 8 Somapah Road, 487372 Singapore
}
\author{Tianqi Chen}
\affiliation{
    School of Physical and Mathematical Sciences, Nanyang Technological University, 639798 Singapore
}
\author{Erping Li}
\affiliation{
    College of Information Science and Electronic Engineering, Zhejiang University, Hangzhou 310027, China
}
\author{Bo Yang}
\affiliation{
    School of Physical and Mathematical Sciences, Nanyang Technological University, 639798 Singapore
}
\author{L. K. Ang}
\affiliation{
    Science, Mathematics and Technology, Singapore University of Technology and Design, 8 Somapah Road, 487372 Singapore
}

%
%

\begin{abstract}
     The tensor network, as a facterization of tensors, aims at performing the operations that are common for normal tensors, such as addition, contraction and stacking. However, due to its non-unique network structure, only the tensor network contraction is so far well defined. In this paper, we propose a mathematically rigorous definition for the tensor network stack approach, that compress a large amount of tensor networks into a single one without changing their structures and configurations.
     We illustrate the main ideas with the matrix product states based machine learning as an example. Our results are compared with the \emph{for} loop and the \emph{efficient coding}  method on both CPU and GPU.
\end{abstract}

\maketitle

\section{Introduction}

Tensors are multi-dimensional arrays describing multi-linear relationships in a vector space. The linear nature of the tensors implies that two tensors may be added, multiplied, or stacked. The stack operation of tensors place all the tensors with the same shape row by row, resulting in a higher-dimensional tensor with an extra stack dimension. For example, stacking scalars will lead to a vector and stacking vectors gives a matrix. The tensor operation is designed for the fundamental calculation in modern chips such as graphics processing unit (GPU) and tensor processing unit (TPU) \cite{TPUGoogle}. Thus, computations with a combination of tensor operations would be more efficient as compared to the conventional ways. For example, there are two ways to calculate the dot product between a hundred vectors $v^{i}_{n\times1}(i=1,2,\cdots,100)$ and a core vector $c_{n\times 1}$: either by applying the dot operation $100$ times via a \emph{for} loop or align all the vectors into a big matrix $M_{100\times n}$ and apply a matrix-vector dot product. The later method is much faster in GPU or TPU due to their intrinsic design.

The tensor network \cite{VerstraeteCirac2008,Orus2014,orus2019tensor,CiracVerstraete2021} is a decomposition of a very large tensor into a network structure of smaller tensors, which has seen recent applications in machine learning\cite{StoudemireSchwab2016,GaoDuan2017,DengDasSarma2017,Stoudenmire2018,ChenXiang2018,HanZhang2018,glasser2019probabilistic,Liu2019,efthymiou2019tensornetwork,roberts2019tensornetwork}. As it is merely the decomposition of the tensors, the tensor networks were expected to be able to maintain the same operations as mentioned previously. However, due to its arbitrary decomposition into a network structure, only the contraction operation is well-defined. Other operations such as `stack' and the batched tensor networks have not been studied thoroughly so far. In fact, the stack operation is quite an important part of the modern machine learning tasks, by treating hundreds of input tensors as a single higher-dimensional \textit{batched} tensor {at training and inference steps}.
Recent tensor network {machine learning approaches} includes
the matrix product states (MPS) \cite{StoudemireSchwab2016,HanZhang2018},
the string bond states (SBS) \cite{glasser2019probabilistic},
projected entangled pair states (PEPS) \cite{ChengZhang2021,vieijra2022generative},
the tree tensor network (TTN) \cite{Stoudenmire2018,Liu2019,ChengZhang2019}
and others\cite{NovikovVetrov2015,GuoPoletti2018} that either use a naïve iteration loop, or skip the batched tensor network via a programming trick (we note it as the efficient coding). {Several studies have explored the concept of stack operation of tensor network to a certain extent. For instance, Ref.~\cite{Weichselbaum2012} puts forward the `A-tensor' to represent the new local unit of MPS after adding the extra local state space. It represents the local tensor of the symmetric tensor network, which is regarded as the stack of different `multiplets'. This approach is developed for the scenario that the system has certain symmetries so that the computational cost can be well controlled. It did not explicitly reveal the concept of stacking hundreds of tensor networks into one network. A more recent work \cite{Ran2022} has put forward the `add' operation of MPS for solving the many-body Sch\"{o}dinger's equation. It represents the coefficients of the many-body wave function as tensor networks and `add' them into one compact form. The `add' operation is similar to our definition of the stacked tensor network with an additional reshape.} Therefore, it is useful to discuss about what is the stack of tensor network and how it performs in machine learning tasks. It is expected that the batched tensor network method may be an alternative option for a faster machine learning program when efficient coding is unavailable. It can also provide the theoretical background for future chips designs based on tensor network contractions.

In this paper, a regular tensor network stack operation method is put forward to compress two or more tensor networks with the same structure into one tensor network with one more stack bond dimension. The remaining part of the stacked tensor network is the same as that of the original tensors.
We illustrate the main procedures with the realization of the stack operation on a one-dimensional tensor network system, i.e., the MPS. However, our result can be extended to arbitrarily shaped tensor network such as the PEPS.
The rest of the paper is organized as follows.
In Sec.~\ref{sec:methods}, we present the concept of the stack operation for tensor networks, as well as the proof using MPS as an example. In Sec.~\ref{sec:Application}, we benchmark the speed and compare the machine learning performance for three batch computing methods for the MPS machine learning task: the naive loop method in Sec.~\ref{subsec:for}, the batch tensor network in Sec.~\ref{subsec:batch}, and the efficient coding method in Sec.~\ref{subsec:efficientcoding}. In Sec.~\ref{subsec:resultdiscussion}, we further discuss about the relationship between the batch tensor network and the efficient coding method. We draw our conclusion and provide future perspective in Sec.~\ref{sec:conclusion}.

\section{Batch contraction of tensor networks}\label{sec:methods}
  In this section, we provide the {definition of the regular} stack operation of tensor networks used in the paper.
  We will first discuss some basic concepts of tensor in Sec~\ref{subsec:Tensor} and tensor networks in  Sec~\ref{subsec:TN}.
  We then present the details of the stack operation for MPS in Sec.~\ref{subsec:stackoperation} and its generalized formation about higher dimensional tensor network in  Sec.~\ref{subsec:stackoperationhigh}.

   \subsection{The stack operation}
   \label{subsec:Tensor}
    A tensor $\mathcal{T}$ is a specifically organized high-dimensional collections of real or complex numbers. For example, a rank-$k$ tensor has $k$ indices. Therefore, a rank-$1$ tensor has only one index, and it is just a vector; a rank-$3$ tensor is a higher-order tensor with three independent indices.
    In this paper, we will use a tuple {with parentheses} $(L_1,L_2,\dots,L_K)$ to represent the rank-{$K$} tensor
    $$
    \mathcal{T} \equiv \mathcal{T}_{p_1,p_2,\dots,p_K}=
    (L_1,L_2,\dots,L_K).
    $$
    {where $p_i$ is the $i^{\text{th}}$ local index which takes the value from $1$, $2$ to $L_i$, i.e. for each $i$, the local dimension is $L_i$.}
    Therefore, a vector which has $N$ dimension could be represented as a tuple like $(N,)$ and a matrix whose size is $N\times M$  is $(N,M)$.

\begin{figure}[h]
        \centering
        \includegraphics[width=.78\columnwidth,draft=false]{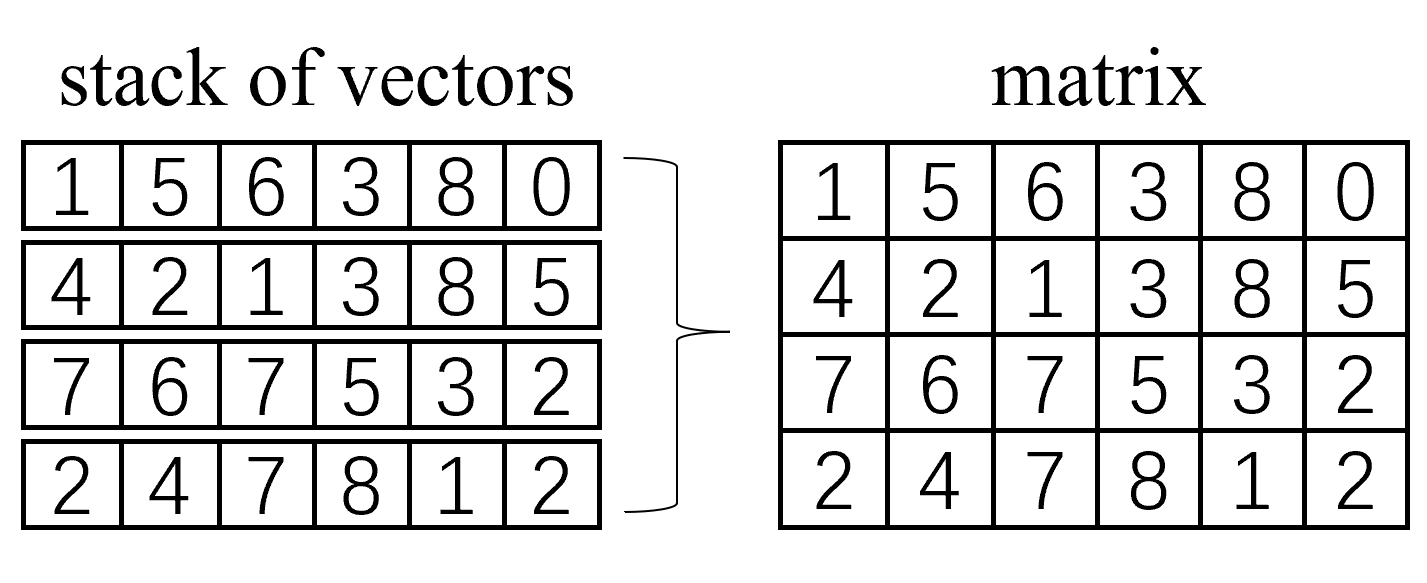}
        \caption{\label{fig:stack_tensors}Illustration of stacking $4$ vectors of $(6,)$ into a single matrix $(4,6)$.}
    \end{figure}

    `Stack' is a pervasive operation in tensor analysis.
    It could merge a series of tensors having the same shape into a higher rank tensor. For example,
    if we stack $N$ vectors $(M,)$, then we get a $(N,M)$.
    If we stack $K$ matrix $(N,M)$, then we get a rank-$3$ tensor $(K,N,M)$ [Fig.~\ref{fig:stack_tensors}].

     In computer science, the stack operation generally converts the list of tensors having the same shape into a very compact form that is memory efficient. For example in Matlab, for a certain problem which requires going through all the elements, transforming it into matrices production is much more efficient than using a for-loop operation for the multiplication of row vectors.

     For modern machine learning or deep learning, the feeding {input} data is always encoded as tensors. In order to deal with thousands of inputs, those tensors are stacked together and reformulated into a higher-rank tensor called \emph{batch} before passing it into the model. This type of higher-rank tensor is also {referred to as \textit{stack} operation in machine learning}.

   \subsection{Basics of tensor networks}\label{subsec:TN}
        Tensor network $\{T^i\}$ is the graph factorization of a very large tensors {by some decomposition processes like SVD or QR. The correspondingly tensor results from contracting all the auxiliary bonds of the tensor network}.
        $$
        \mathcal{T} \xrightleftharpoons[\text{contraction}]{\text{decomposition}} \{T^i\}
        $$
        where $T_i$ represent the small local tensor in tensor network.

\begin{figure}[h]
           \centering
           \includegraphics[width=1.0\columnwidth,draft=false]{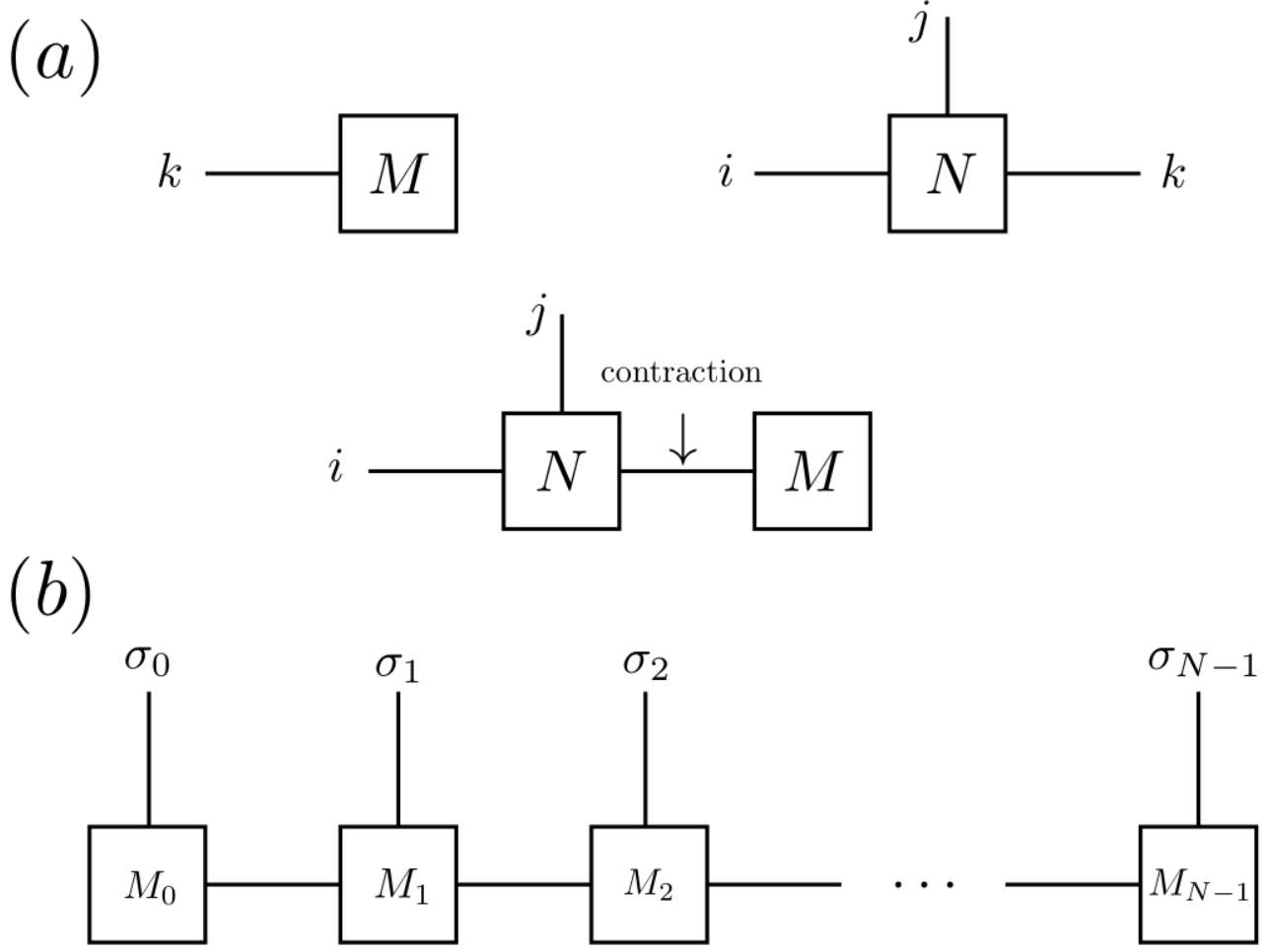}
           \caption{\label{fig:TN} Diagrammatic representations for tensor networks: (a) Diagrammatic representations for tensors $M_i$ and $N_{i,j,k}$, and the contraction between them $M_i \cdot N_{i,j,k}$; (b) Diagrammatic representation for one-dimensional tensor network MPS: $M_j$ is the local rank-$3$ tensor and each is contracted via auxiliary bond dimensions (vertical solid lines); $\sigma_j$ is the index for the physical dimension.}
       \end{figure}

    {A graphical representation of tensors is shown in Fig.~\ref{fig:TN}(a).}
       In general, a rank-$k$ tensor has $k$ indices.
       Therefore, a vector as the rank-$1$ has one dangling leg,
       and a rank-$3$ tensor has three dangling legs, as shown in Fig.~\ref{fig:TN}(a).
       The contraction between two tensors can thus be defined by contracting the same index between two arbitrary tensors. In Fig.~\ref{fig:TN}(a), a rank-$1$ tensor $M_k$ is contracted with another rank-$3$ tensor $N_{i,j,k}$ as $M_k\cdot N_{i,j,k}$, and it ends up with a rank-$2$ tensor, i.e. a matrix.

       In the context of quantum physics, a many-body wave function can be represented as a one-dimensional rank-$N$ tensor network where each leg represents the physical index for each site, as shown graphically in Fig.~\ref{fig:TN}(b). This is usually referred to as the matrix product state (MPS). The MPS can therefore be seen as a connected array of local rank-$3$ tensors. Each tensor $M_j$ has three indices: the physical index $\sigma_j$, and two auxiliary indices $D_j$ and $D_{j+1}$ that are contracted and therefore implicit \footnote{Note that for the first and the last site, one of the auxiliary indices is a dummy index and therefore can be neglected graphically.}. An MPS with open boundary condition can be written as
       \begin{align}
            & \ket{\psi}=\sum_{\{\bm{\sigma}\}}\left[M_0^{\sigma_0}M_1^{\sigma_1}\dots M_{n-1}^{\sigma_{n-1}}\right]|\sigma_0,\dots,\sigma_{n-1}\rangle
       \end{align}

   \subsection{Stack operation for MPS}\label{subsec:stackoperation}

  A tensor $\mathcal{T}$ can be converted to a tensor network $\{T^i\}$ which consists of contracted local tensors $T_i$ via decomposition such as the tensor train method \cite{oseledets2011tensor}. {Mathematically, one could perform operations such as product, trace, contraction, splitting and grouping, etc. on both of them.}

\begin{figure}[h]
           \centering
           \includegraphics[width=1.0\columnwidth,draft=false]{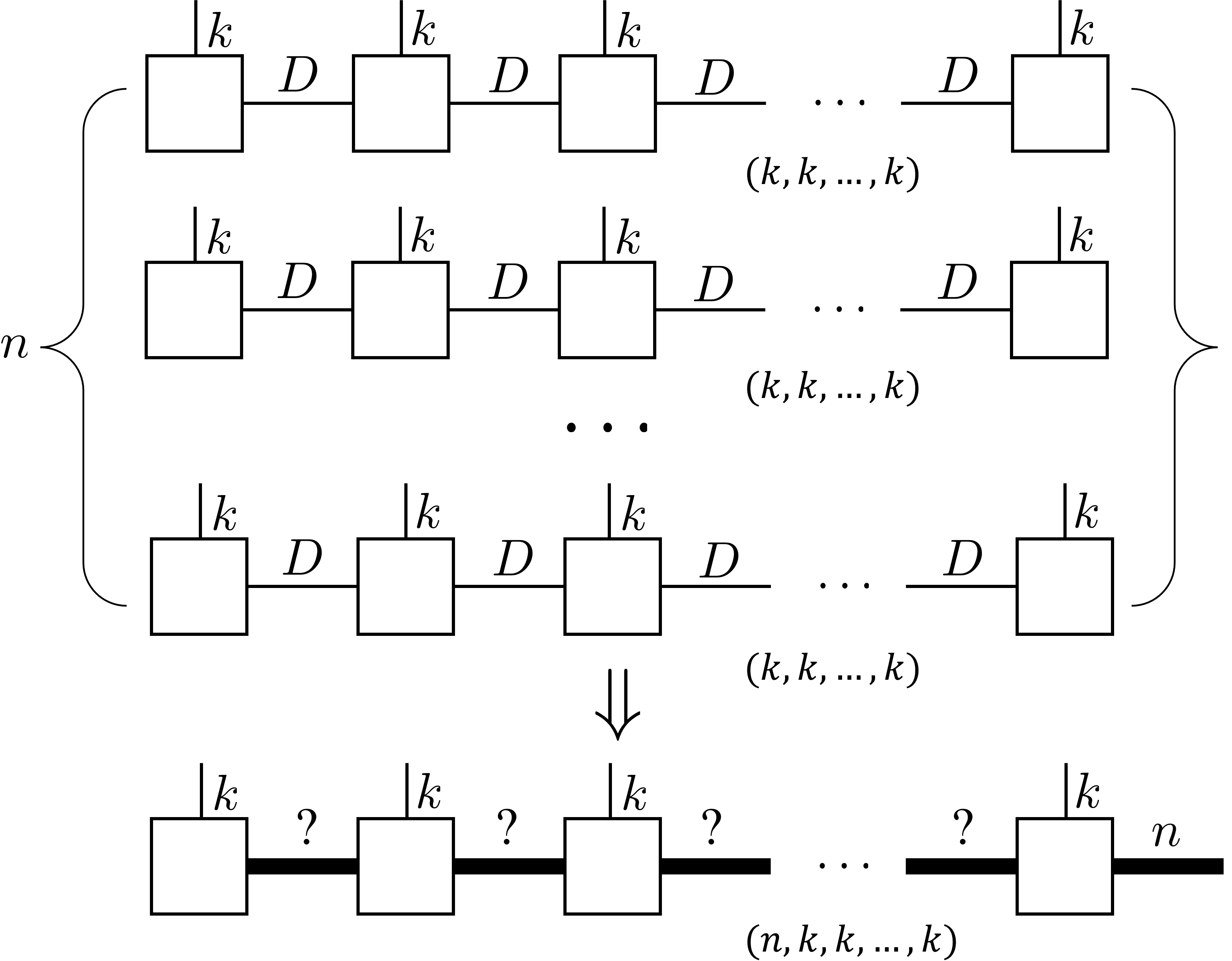}
           \caption{\label{fig:diagram_stack}
           Schematics of stack operations for MPS: Illustration for the $n$ identical MPS with auxiliary bond dimension $D$ which ends up with a new MPS with same physics dimension. One of sites should has an extra bond dimension $n$ to fulfill the stack dimension requirement. (Here we set the last site)
           }
       \end{figure}

  However, the stack operation can only be directly applied to one single tensor network rather than to multiple ones separately. More precisely, we want to deal with such a problem as shown in Fig.~\ref{fig:diagram_stack}:
  given a series of tensor networks $\{T^i\}_{\alpha}$ with the same configuration, we want to get their stacked tensor network representation with the same physical structure and shape.

       One direct method to realize this is to convert those tensor network $\{T^i\}$ firstly to the corresponding tensor $(k,k,\dots,k)$,
      and then decompose the stacked tensor  $(n,k,k,\dots,k)$.
      However, this deviates from the original intention of representing a tensor as a tensor network. Note that we use a tensor network to avoid storing or expressing the full tensor in computing, which is memory-consuming and may be prohibited in large systems.
      Thus we would like to explore other approach that can help us efficiently to establish the representation of the stacked tensor network without accessing any contraction.

\begin{figure}[h]
          \centering
          \includegraphics[width=.8\columnwidth,draft=false]{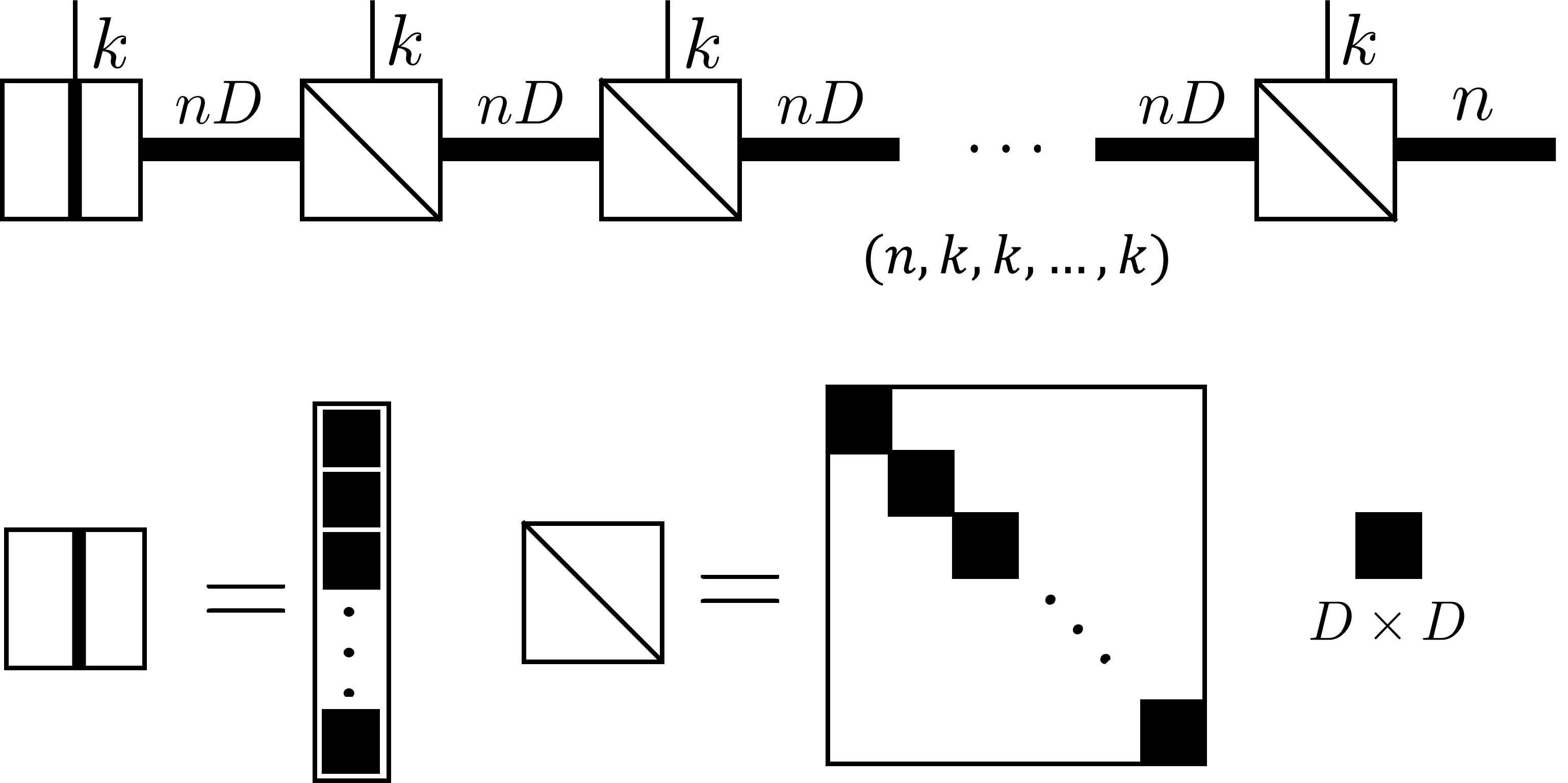}
          \caption{\label{fig:stackoperation_result} The diagram result shows the result of the `stack' operation for MPS. It will allocate a block diagonal local rank-$3$ tensor:
          for each physical dimension, it gives a block diagonal matrix where the size of each block is a matrix $(D,D)$. Note the first tensor unit's block is a vector $(D,)$ and the last unit's block is a matrix $(D,1)$ }
      \end{figure}

We show in Fig.~\ref{fig:stackoperation_result}: each rank-$3$ tensor in MPS is $(k,nD,nD)$.
      Each matrix $(nD,nD)$ along physical dimension $k$, is a block diagonal matrix.
      The $i$-th block $(D,D)$ is exactly the $i$-th rank-$3$ tensor in stack list at the same physical dimension $k$ (so it is a matrix).
      There are minor differences between the start and the end of the MPS. The start tensor unit is a rank-$2$ matrix with a row stack for the first unit in stack list.
      The end tensor is a diagonal rank-$3$ tensor $(k,nD,n)$ with each diagonal $D \times 1$ block filled by the end unit in stack list.
      A full mathematical proof is given in Appendix.~\ref{app:mathproof}.

   \subsection{Batch operation for general tensor networks}\label{subsec:stackoperationhigh}

           For other types of tensor network like PEPS, the unit tensor is a $4+1$ rank tensor. The formula shown in \ref{subsec:stackoperation} would be more complicated.
           We post the results here directly:
              For any tensor network consisting of $M$ tensor unit
              $$
              \{T^{i}=(k,L_1,L_2,\dots,L_m)|i=1,2,\dots,M\}
              $$
              its $n$-batched tensor network version is another tensor network
              $$
              \{T^{i}_{\text{batch}}=(k,nL_1,nL_2,\dots,nL_m)|i=1,2,\dots,M\}
              $$
              with one extra  {The extra index represents the stack dimension(corresponding to the tensor stack) is free to add since each local tensor is block-diagonal. We only need to unsqueeze/reshape the diagonal block from $(k,L_1,L_2,\dots)$ to $(k,1,L_1,L_2,\dots)$. For the MPS case,  we could put it at the right end, as shown in Fig.~\ref{fig:stackoperation_result}.}

              The sub-tensor $(nL_1,nL_2,\dots,nL_m)$ along the physical indices
              for each tensor is a diagonal block tensor.
              For example, the $j$-th diagonal block is
              $$
              (T^{i}_{\text{batch}})_{k,p_1,p_2,\dots,p_m} = T^{i}_j
              $$
              where $p_i$ is the local index for index $i$ which takes the value from $p_{i}\in[j*L_{i},(j+1)*L_{i}]$ {and $T^i_j$ is the $T_i$ of the $j$-th tensor network in tensor network list $\{T_i\}_\alpha$.}

              Meanwhile, one unit (for example, take the $N$-th local tensor) should be `unsqueezed' to generate the dangling leg for the stack number indices
              $$
              {T}^{N}_{batch} = (k,nL_1,nL_2,\dots,nL_m,n).
              $$
              For example, the MPS case in \ref{subsec:stackoperation}, will generate a row vector which is just the `diagonal block' form for the  rank-$1$ tensor.
              And the last MPS unit would get an external `leg' with dimension $N$ is the requirement of the batch.

\section{Application}
\label{sec:Application}
    In this section, we study the potential applications of the stack of tensor network. We apply this stack operation method to the tensor network machine learning.
    For a tensor network based machine learning task,
    both the model and input data can be collectively represented as a tensor or tensor network.
    {In fact, here, the model actually refers to the trainable parameters, and we would note it as machine learning core in the following context. The forward processing to calculate a response signal $y$ corresponding to the input $x$ is an inner product in a linear or nonlinear function space, as the interaction between the machine learning core and the input data. The lost function is designed based on the type of task. For example, we would measure the distance between the real signal $\hat{y}$  and the calculated signal $y$; for unsupervised learning, we might design the structure loss of hundreds of signals $y$ to maximize the distance matrix. In this work, we take supervised machine learning as an example. However, such a method is flexible for any machine learning framework since it is a modeling technology. }
    Here, by using the language from quantum physics, we denote the machine learning core as $| \mathcal{C} \rangle$ and one of the input data as $|I_j=I_j(\psi) \rangle$. As both the core $| \mathcal{C} \rangle$ and the input $|I_j \rangle$ are represented as tensors, this processing can be batch executed in parallel.
    This machine learning task can then be efficiently represented as
    \begin{align}
        \label{eq:MLEq1}
         & \langle \mathcal{C} |(B,I_j)\rangle = (B,\langle \mathcal{C} |I_j\rangle)=(B,\alpha_j)
    \end{align}
    where $\alpha_j$ is the resulting output data, and  $\ket{(B,I_j)}$ is the batch representation of $B$ stacked tensors or a tensor network, where $\ket{(B,I_j)}$ has an explicit format.

\begin{figure}[h]
        \centering
        \includegraphics[width=1.0\columnwidth,draft=false]{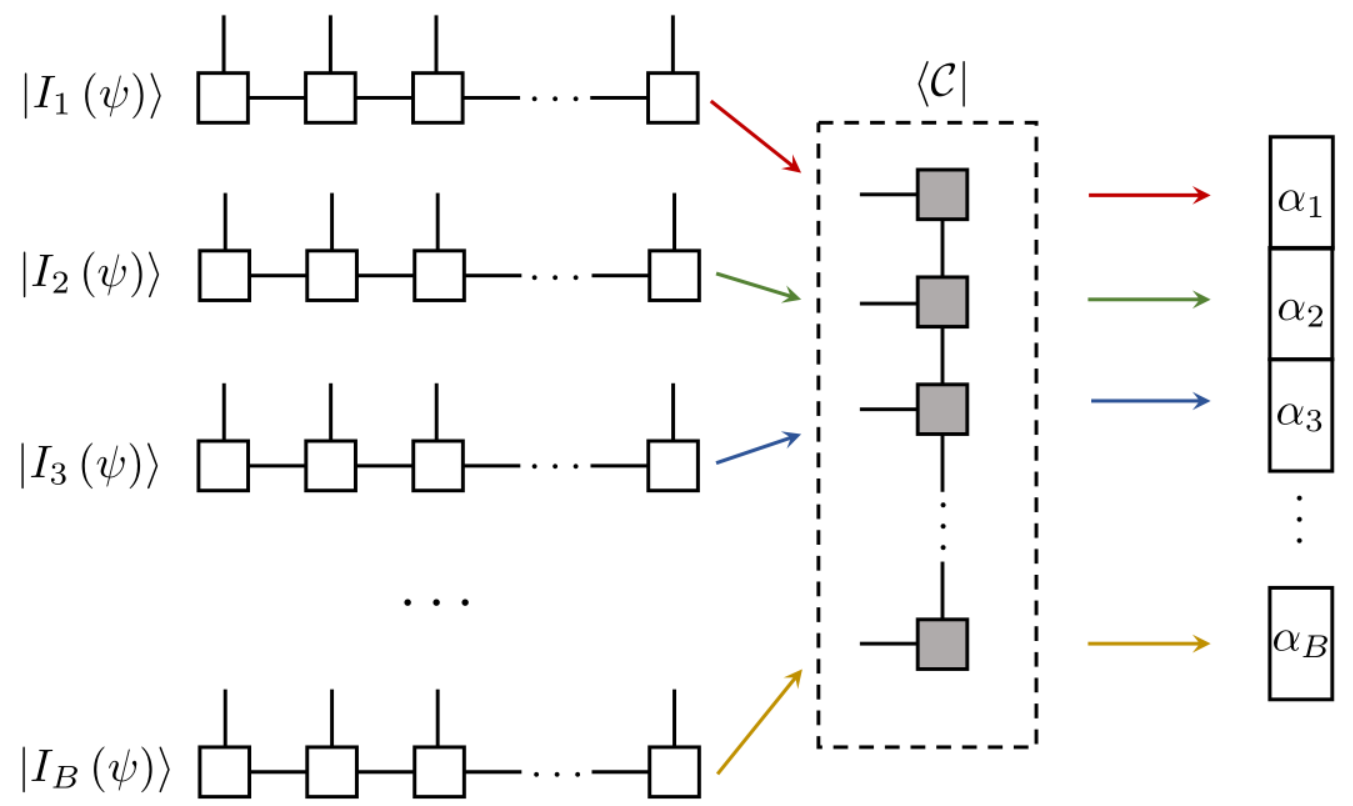}
        \caption{\label{fig:stacktaskML} Schematics of inputting individual data as MPS ($\ket{I_j\left(\psi\right)}$) onto the core model ($\bra{\mathcal{C}}$): each operation follows the arrow with the same color, and results in an outcome $\alpha_j$.}
    \end{figure}

    In the following, we consider the one-dimensional case of the MPS machine learning task as an example.
    As shown in Fig.~\ref{fig:stacktaskML}, both the core $| \mathcal{C} \rangle$ and the inputs $|I_j \rangle$  are the MPS, and the output responsive signal $\alpha_j$ is the result obtained by contracting both physical bonds from the two MPS.

    Below we show that there are three ways to accomplish this machine learning task: the loop (LP) method,
    the batched tensor network (BTN) method and the efficient coding (EC) method. The core MPS $| \mathcal{C} \rangle$ consist of $L$ local tensor units with $k$ being the dimension for all physical bonds and $V$ being the dimension for all auxiliary bonds. Each input MPS $|I_j \rangle$ consists of $L$ local tensor units with $k$ being the dimension for all physical bonds and $D$ being the dimension for all auxiliary bonds.

    \subsection{Na\"ive loop method}\label{subsec:for}
      As shown in Fig.~\ref{fig:stacktaskML}, for $B$ {input samples}, one needs to perform $B$ contractions to obtain $B$ outcomes.
      The most obvious way to accomplish this is to perform contractions one by one. When we utilize the auto differential feature from the libraries \textsf{PyTorch} \cite{pytorchref} and \textsf{Tensorflow} \cite{tensorflow2015-whitepaper}, the gradient information would be accumulated in every step. At the end of the loop, we can acquire the batch-gradient by averaging over the sum, and apply it to the gradient descent method for weight updating.

    \subsection{Batched Tensor Network Method}\label{subsec:batch}
\begin{figure}[h]
             \centering
             \includegraphics[width=.8\columnwidth,draft=false]{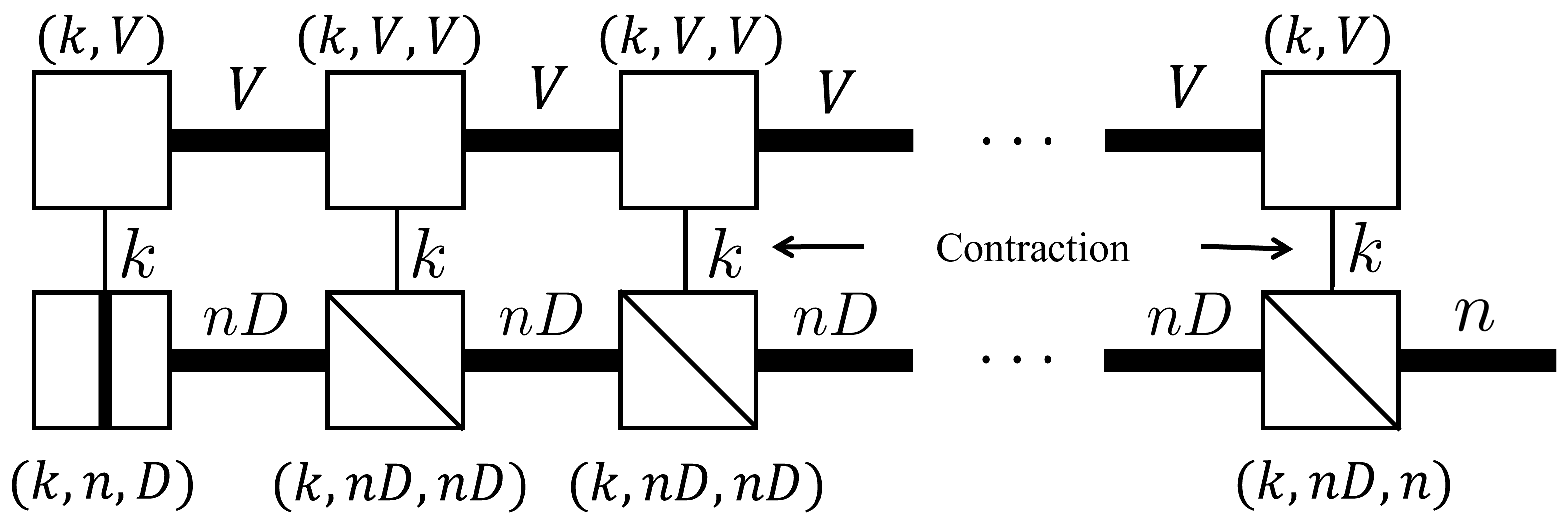}
             \caption{\label{fig:batched_contraction_diagram} Schematics of contraction between the core MPS $\bra{\mathcal{C}}$ (upper), and the batched MPS $\ket{(B,I_j)}$ (down).}
         \end{figure}

        As discussed in Sec.~\ref{sec:methods}, we can stack all the inputs MPS to a batched tensor network which behaves like another MPS. $\ket{(B,I_j)} = \ket{I(\psi_{\rm stack})}$.
        Assume the number of input is $B$, although the auxiliary bond dimension of the batched MPS  $\ket{(B,I_j)}$ gets $B$ times larger [see Fig.~\ref{fig:stackoperation_result}],
        the action of the inner product
        \begin{align}
            \label{eq:MLEq}
             & \langle \mathcal{C} |(B,I_j)\rangle = \langle \mathcal{C} \ket{I(\psi_{\rm stack})}
        \end{align}
         can now be written as one single tensor contraction, as shown in Fig.~\ref{fig:batched_contraction_diagram}
         .
         With the help of the format of the batched tensor network, we can now compress $B$ times contraction into one single contraction.
         Our advantage is obvious: software libraries such as \textsf{PyTorch} and computational units such as GPU are intrinsically optimized for tensor operation, so contracting a large tensor is usually much faster than the matrix multiplication using a for loop.
         As you can see in the benchmark shown in Fig.~\ref{fig:benchmark}, when using the GPU environment, the BTN method can get constant time complexity while the LP method is the linear function of batch size.

         Notice that we need to calculate an optimized contraction path to efficiently contract the tensor network as shown in Fig.~\ref{fig:batched_contraction_diagram} while the LP method can simply contract physical bond first and then sweep from left to right. When $B$ gets larger, contracting physical bond first in order requires storing an MPS with auxiliary bond equal to $nVB$, which is quite memory-consuming. Such a method would cost extra memory allocation for the block diagonal tensor. The memory increment is $O(B^m)$, where $m$ is the number of the auxiliary bond. For example, the $rank$-3 tensor of the batched MPS will allocate a $k\times nB \times nB$ memory to store only $B\times k\times n\times n$ valid parameter.
         If an autodifferential action is needed in the following process, the intermediate tensor needed for gradient calculation will also become $B$ times larger.
         Thus, although the speed is faster when the batch goes larger, the memory will gradually blow up. The BTN method is a typical `memory swap speed' method.

          One possible solution is to treat the diagonal block tensor as a sparse tensor, so we only need to store the valid indices and values.
          Meanwhile, if we contract the physical bonds first, it will return a matrix chain whose unit is also a sparse diagonal matrix of the shape $(nBV,nBV)$.
          This implies that we can block-wise compute the matrix chain, so that each sparse unit performs like a compact tensor $(B,nV,nV)$. At this juncture, it turns out that we can then use the popular coding method called efficient coding, which is widely used in modern tensor network machine learning task \cite{HanZhang2018,
            ChengZhang2021,
            ChengZhang2019,
            StoudemireSchwab2016,
            efthymiou2019tensornetwork}.

   \subsection{Efficient coding}\label{subsec:efficientcoding}
      The efficient coding (EC) does the `local batch contraction' for every tensor unit calculation in the tensor network contraction.

\begin{figure}[h]
          \centering
          \includegraphics[width=.8\columnwidth,draft=false]{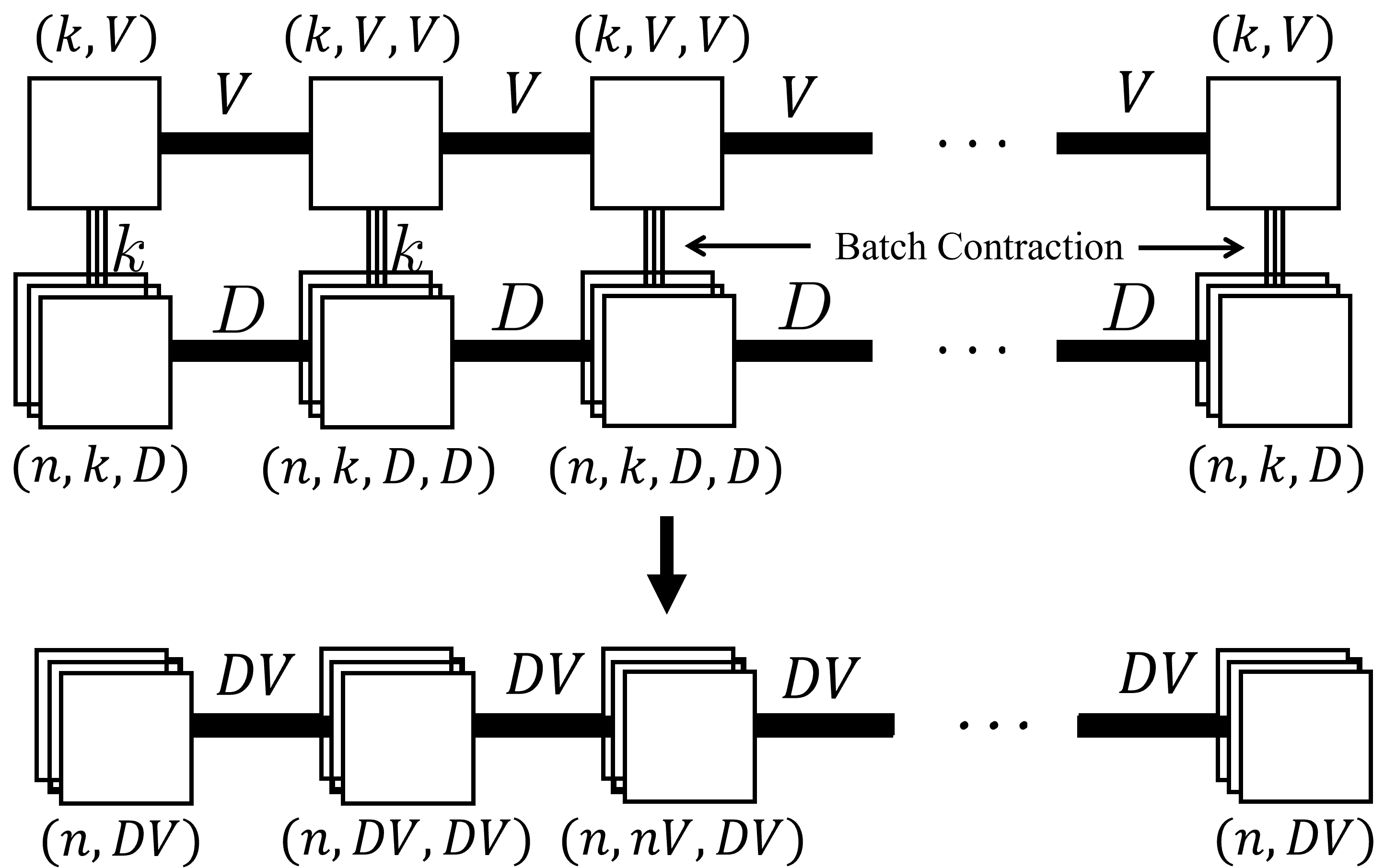}
          \caption{\label{fig:efficient_coding} Illustration of efficient coding for MPS. Each unit from the core MPS is batch contracted with input MPS (upper), and they form a batched unit as a local tensor in the resulting MPS (down).}
      \end{figure}

      We first batch contract each physical index and get a tensor train consist of batched units as shown in Fig.~\ref{fig:efficient_coding}, then we perform the batch contraction on all the auxiliary bonds for this batched tensor train.

      The batch contraction is realized by the built-in function \textsf{einsim} which is provided in many modern scientific software libraries such as \textsf{Numpy} \cite{harris2020array}, \textsf{PyTorch} \cite{pytorchref} and \textsf{Tensorflow} \cite{tensorflow2015-whitepaper}:
      \begin{align}
      &\text{einsum('kij,Bknm} \rightarrow \text{Binjm',core,batch)}\\
      &\text{einsum('Bij,Bjk}  \rightarrow \text{Bik',core,batch)}
      \end{align}
      For those  without \textsf{einsim}, one alternative solution is to implement a highly efficient `batch matrix multiply' function. The fundamental realization of \textsf{PyTorch.einsim} is in fact the `batch matrix multiply' called \textsf{PyTorch.bmm}.
      {Notice such a method is not new for tensor network machine learning, and Refs.~\cite{HanZhang2018,efthymiou2019tensornetwork,ChengZhang2021} has already realized successful learning algorithms based on this.}

      The memory allocated for EC is much less than that for the BTN, which is the major reason why the BTN method is slower than EC.

      Also, the dense BTN may require an optimized contraction path for a large batch case, while EC for MPS would only allow contraction of the physics bond first, followed by the contraction for the auxiliary bonds as the tensor train contraction. The standard tensor train contraction method sweeps from the left-hand side of the MPS to the right.
      In particular, when we require all MPS in the uniform shape, i.e., all the tensor units share the same auxiliary bond dimension, we can then stack those units together($(B,L,V,V)$) and split it by odd indices ($(B,L/2,V,V)$) or even indices ($(B,L/2,V,V)$).
      In doing so, we can get the half-length tensor units ($(B,L/2,V,V)$) for the next iteration by contracting the odd part and even part.

    \subsection{Comparison and discussion}\label{subsec:resultdiscussion}
     \begin{figure*}[ht]
         \centering
         \includegraphics[width=2.0\columnwidth,draft=false]{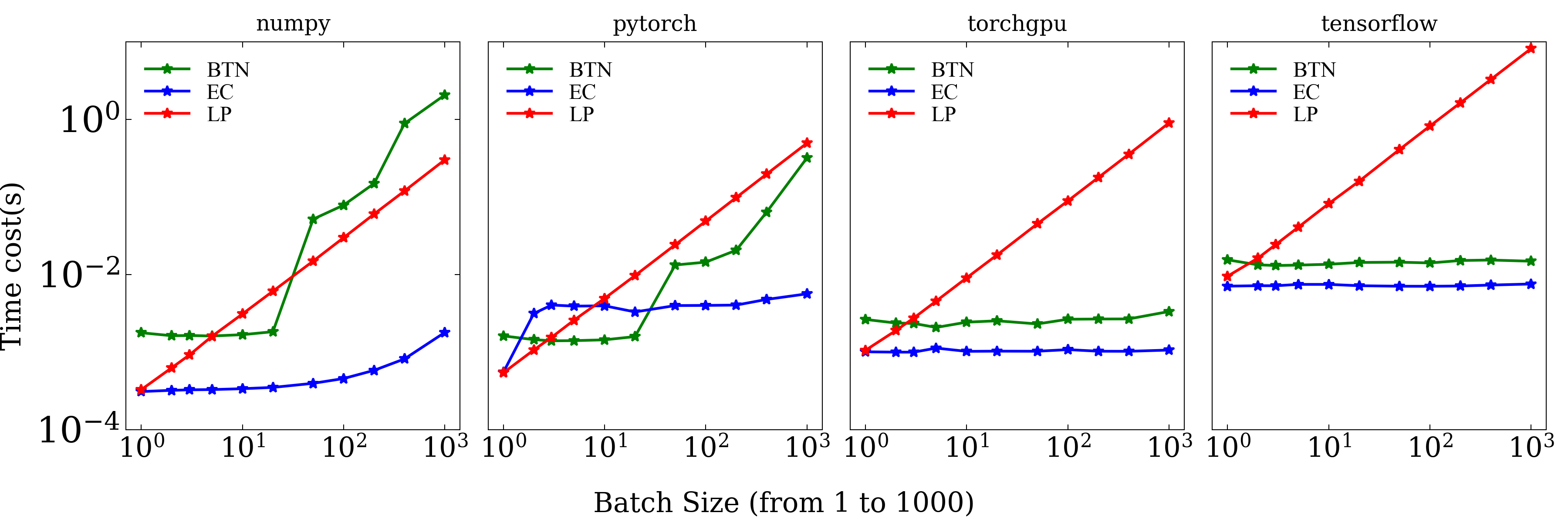}
         \caption{\label{fig:benchmark} Speed-up benchmark results for different methods as a function of batch size with: (a) Numpy on CPU; (b) PyTorch on CPU; (c) PyTorch on GPU; (d) Tensorflow on GPU. The red, blue and green solid lines correspond to the resulting obtained using the loop method (LP), efficient coding (EC), and the batch tensor network method (BTN) respectively. All the panels are plotted in $\log-\log$ format. All the result is test on a Intel(R) Core(TM) i7-8700K CPU and 1080Ti GPU.}
     \end{figure*}
     In Fig.~\ref{fig:benchmark}, we show the speed-up benchmark results for all three different methods introduced from Sec.~\ref{subsec:for} to Sec.~\ref{subsec:efficientcoding} with respect to different batch size in a $\log-\log$ plot.
     We compare the performance using different libraries in \textsf{Python}.
     We mainly test three batch contraction methods:
     the Loop method (LP),
     the batch tensor network method (BTN), and
     the efficient coding method (EC) on the MPS batch contraction task in Fig.~\ref{fig:stacktaskML}. The system size is a $L=20$ units MPS with each unit assigned $V=6$, $k=3$ and $D=1$ [see Fig.~\ref{fig:efficient_coding}].
     Batch contraction task size from $1$ to $1000$.
     Overall, EC has the best speed-up performance over other methods for almost all libraries, especially at large batch size.
     Due to the pre-processing requirement, the BTN can be slower than others in small-batch cases.
     The CPU is not efficient in storing and processing larger matrices or tensors, while the GPU is deeply optimized to deal with tensor data structure.
     Thus, the BTN gets similarly linear time complexity like LP in CPU but maintain a constant time complexity in GPU.
      Note the BTN will allocate tensors with $(nB,nB)$. Thus the larger the batch size, the larger the intermediate tensor. We can see that the \textsf{PyTorch}(CPU) gets better optimization with large tensor than \textsf{NumPy}.
     In Fig.~\ref{fig:benchmark}(c) and (d), both BTN and EC have constant time complexity in GPU since the GPU would compress the  time complexity of matrix contraction to a constant level. It also imply that there is a close relationship between EC and BTN: As we discussed at the end of the Sec.~\ref{subsec:batch}, the EC is the another realization of the sparse matrix BTN. The redundancy memory requirement for the diagonal block tensor slows down the BTN method, but such redundancy requirement is eliminated in EC.

\begin{figure}
         \centering
         \includegraphics[width=.8\columnwidth,draft=false]{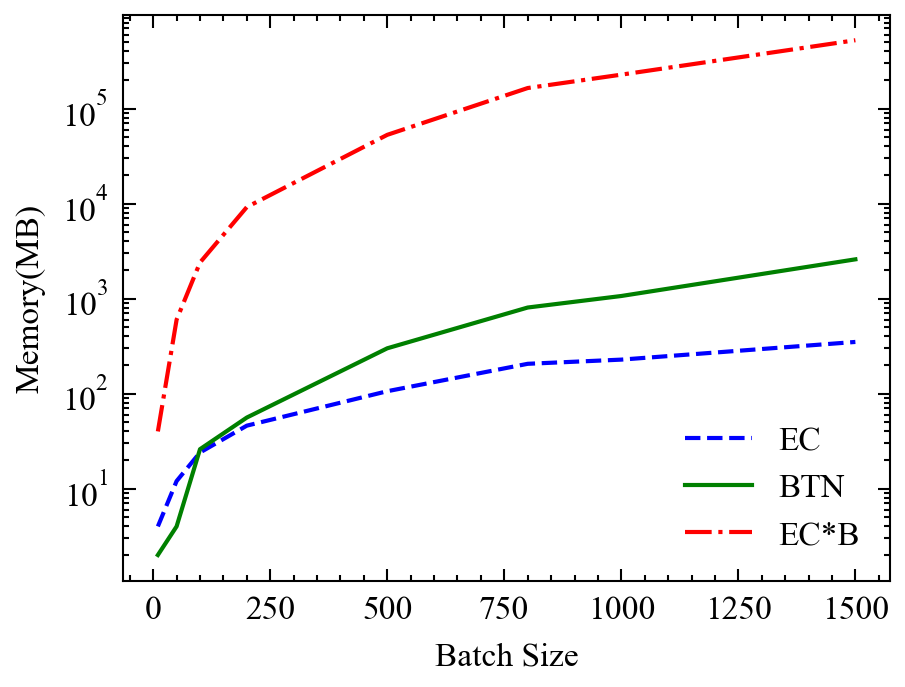}
         \caption{{The memory needed for the increasing batch size MPS machine learning problem. The system is an MPS of $L=21$ units, with each unit assigned with $V=50,k=3,D=1, O=10$. The green curve($BTN$) is the required memory  for BTN method; The blue curve($EC$) is the required memory for EC method; The red curve($EC*B$) is the $B$ times larger curve of $EC$. }}
         \label{memory_batch_modes1}
     \end{figure}

\begin{figure}
         \centering
         \includegraphics[width=.8\columnwidth,draft=false]{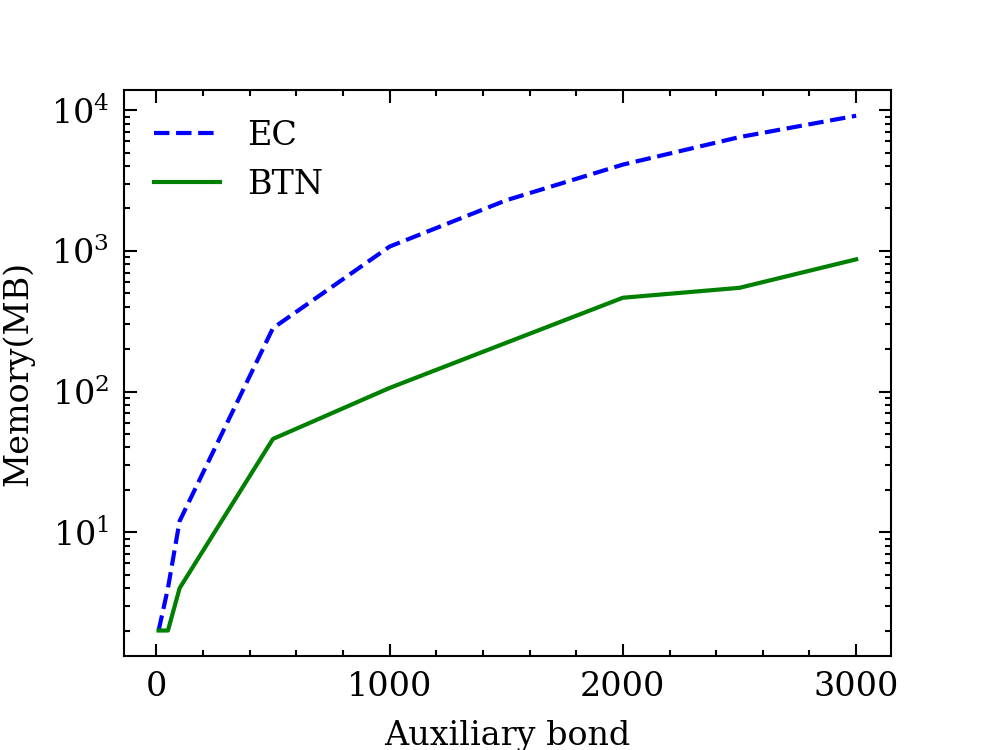}
         \caption{{The memory needed for the increasing auxiliary bond dimension MPS machine learning problem. The system is an MPS with $L=21$ units, with each unit assigned with $B=10,k=3,D=1, O=10$. The green curve($BTN$) is the required memory  for BTN method; The blue curve($EC$) is the required memory for EC method. }}
         \label{memory_batch_modes2}
     \end{figure}

     { For EC and LP method, we contract the physical bond first, and obtain a tensor chain-like structure
            $$
            (B,V)-(B,V,V)-(B,V,V)-\dots-(B,V,V)-(B,V,O)
            $$
        where $B=1$ for the LP method and $B>1$ for the EC method.
        Notice that we assume $D=1$ here for the traditional MPS machine learning task.
        The memory complexity costs $(L-1)\times BV^2+BVO+BV$ as the $O(B)$.
        Since the output bond is at the right-hand side, the optimal path is contracting from left end to the right end.
        Here, we use the auto differential engine to automatically calculate the gradient for each unit, then the computer records all the intermediates during the forward calculation. We therefore obtain the $(B,V)$ batched vector calculated by a batched vector-matrix dot for the first two units
        \begin{align}
        &(B,V)\times(B,V,V)\rightarrow(B,V) \quad \text{for the rest}\notag\\
        &(B,V)\times(B,V,O)\rightarrow(B,O) \quad \text{for the last}\notag
        \end{align}
        So the memory cost for EC and LP method is straightforward as saving $L-1$ times batched vector-matrix dot result, which is $(L-2)\times BV+BO$.
        If we apply the same contraction strategy to the BTN method, the resulted chain is
        $$
        (BV,)-(BV,BV)-\dots-(BV,BV)-(BV,BO).
        $$
        The memory complexity then costs $(L-1)\times B^2V^2+B^2DVO+BV$ memory as the $O(B^2)$.
        Surprisingly, the intermediate cost is the same: $(L-2)\times BV+BO$.
        However, the left-right sweep path is not always the optimal path for the BTN method, as we discussed in Sec.~\ref{subsec:batch}.
        In our experiment, the memory complexity for the optimal path consumes much less than $O(B^2)$. In Fig.~\ref{memory_batch_modes1}, we give an example of the memory cost for increasing batch size which takes a $L=21$ units of MPS with each unit assigned with $V=50,k=3,D=1$ as the system parameters. The output $O=10$ classes leg is set at the right end. The batch size increases from $B=10$ to $B=1500$. We also plot the $B$ times larger curve of the $EC$, and we simply labeled it as $EC*B$. We could see that although the BTN method takes more memory than the EC method, the memory complexity is much less than $O(B^2)$, which makes it an available option for machine learning.
        Also, for other problems which require small batches but larger auxiliary bond dimensions, the BTN will perform better than the EC method as shown in Fig.~\ref{memory_batch_modes2} with a $B=10,k=3, D=1, O=10$ and $V=10\rightarrow 3000$ MPS learning system. Contracting the auxiliary bonds first and keeping comparably smaller intermediates is a better choice when $B<<V$. This is the reason why we see the BTN method can allocate much less memory than the EC method for large auxiliary dimensions in Fig.~\ref{memory_batch_modes2}}

     \begin{figure}
         \centering
         \includegraphics[width=.8\columnwidth,draft=false]{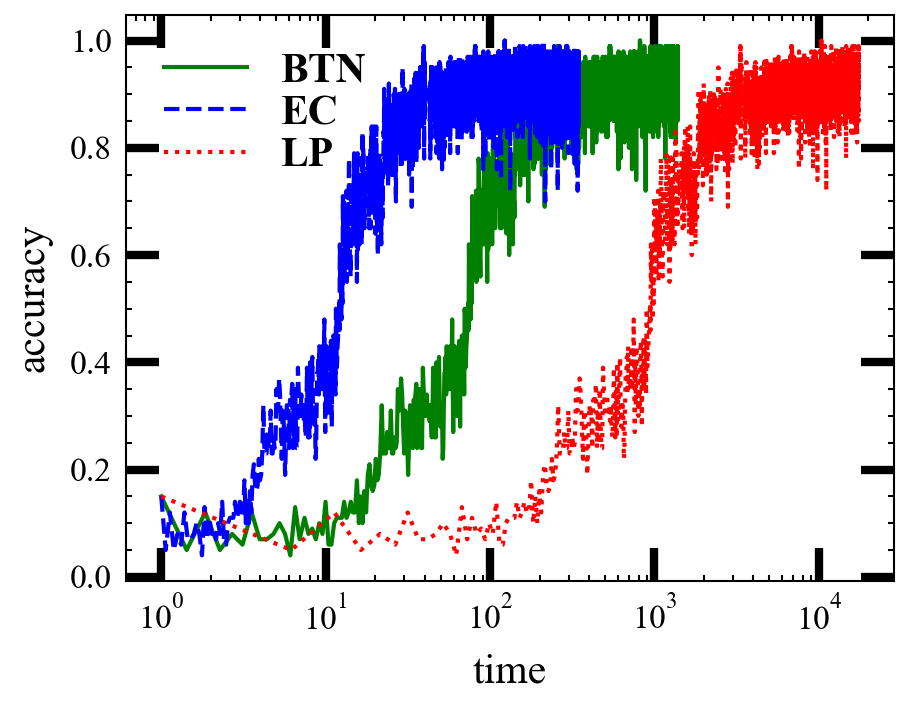}
         \caption{\label{amachinelearningdemo}
         Evaluation of the train accuracy of MPS machine learning tasks on MNIST dataset using three batch contraction method: (red solid curve) the batch tensor network method (BTN); (green dashed curve) the efficient coding; (blue dotted curve) the loop method (LP).}
     \end{figure}
      \begin{figure}
         \centering
         \includegraphics[width=.8\columnwidth,draft=false]{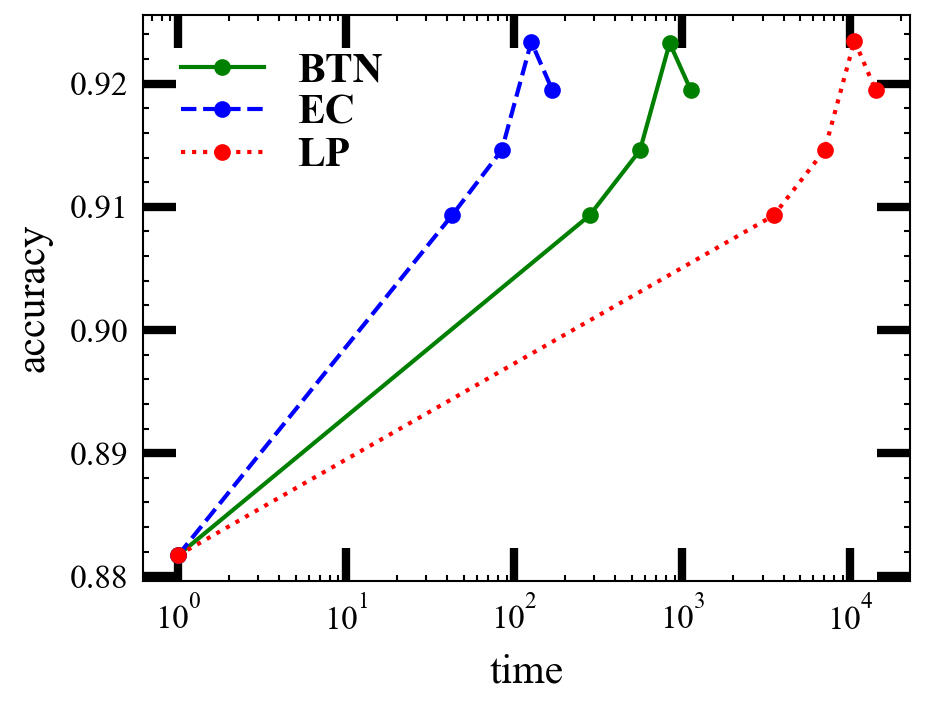}
         \caption{\label{amachinelearningdemovalid}
         Evaluation of the valid accuracy of MPS machine learning tasks on MNIST dataset using three batch contraction method: (red solid curve) the batch tensor network method (BTN); (green dashed curve) the efficient coding; (blue dotted curve) the loop method (LP).}
     \end{figure}

     We also demonstrate a practice MPS machine learning task in Fig.~\ref{amachinelearningdemo} and Fig.~\ref{amachinelearningdemovalid}.
    The dataset we used here is the MNIST handwritten digit database \cite{mnistdata}.
    The machine learning task requires predicting precisely the number from $0$ to $9$ among the dataset.
    The core tensor network is a ring MPS consist of $24 \times 24$ unit with each unit is $(V,k,V)$ except the last one $(C,V,k,V)$.
    Here, $V=20$ is the  bond dimension, $k=2$ is the physics bond dimension, and $C=10$ is the number of classes.
    The  bond dimension of the input tensor network is $D=1$.
    We use the ring MPS as it is easier to train than the open boundary MPS.
    {Fig.~\ref{amachinelearningdemovalid} shows the valid accuracy v.s. time. The valid accuracy is measured after each epoch and only plots $5$ epochs since the model get over-fitting later. Each epoch will swap the whole dataset once.}
    The best valid accuracy for this setup is around 92\%, while the state-of-the-art MPS machine learning task \cite{ChengZhang2021} can reach 99\% with a large  bond  \footnote{The question on how to perform training in MPS machine learning, for example, how to initial the weight and pre-process data, is beyond the scope of this work, and we would refer to Ref.~\cite{HanZhang2018,ChengZhang2021,ChengZhang2019,StoudemireSchwab2016,efthymiou2019tensornetwork}  for reference.}.

      The result is tested on an Intel(R) Core(TM) $i7-8700K$ CPU and $1080Ti$ GPU and is shown in Fig.~\ref{amachinelearningdemo}. The optimizer we used here is Adadelta\cite{zeiler2012adadelta} with the learning rate $\text{\rm lr=}0.001$.
       The batch size is $100$.
       The random seed is $1$.
      The test environment is \textsf{PyTorch}.
      The $x$-axis in Fig.~\ref{amachinelearningdemo} is the time cost. Notice these three methods are the only different approaches to realize the batch contraction so that they won't influence the contraction result.
      That is why these three curves in  Fig.~\ref{amachinelearningdemo} are exactly the same.
      The EC method is the best choice for the tensor network machine learning. The LP method is the most inefficient method, which would cost one hour to traverse the dataset while the EC only needs $60$ seconds and BTN takes $300$ seconds.

\section{Conclusions}\label{sec:conclusion}
    We have investigated the stack realization of the tensor network as the spread concept of the \emph{stack} operation for tensors.
    The resulting stacked/batched tensor network consists of block-diagonal local tensors larger bonds.
    We connected this method to the efficient coding batch contraction technique which is widely used in tensor network machine learning.
    An MPS machine learning task has been used as an example to validate its function and benchmark the performance for different batch sizes,  different numerical libraries, and different chips.
    Our algorithm provides an alternative way of realizing fast tensor network machine learning in GPU and other advanced chips.
    All the code used in this work is available at \cite{TNProject}.

    Several possible future directions may be of significant interest.
    Our work reveals an intrinsic connection between the block-diagonal tensor network units and the batched contraction scheme, which is potentially useful for helping researchers realize faster tensor network implementation with the block-diagonal design like  \cite{vieijra2022generative}.
    Secondly, our algorithm comes without performing SVD to the stacked tensor network; It would be great to define a batch SVD and check its performance and error when truncating the singular values for larger data size.
    Thirdly, the efficient coding method requires the contraction of the physical dimensions first. However, the batch method provides the possibility to start contraction in other dimensions first, which may be useful for the contraction on a two-dimensional PEPS or through tensor renormalization group method \cite{LevinNave2007,GuWen2008,evenblyVidal2015}.
    Finally, the BTN method takes advantage of compressing thousands of contractions into one contracting graph.
    From the hardware design perspective, with a specially designed platform optimized for tensor network contraction, the BTN method may intrinsically accelerate the tensor network machine learning or any task involving a stacked tensor network.
    Meanwhile, the tensor network states in many-body quantum physics has a non-trivial relationship with the quantum circuits. So the idea of representing several tensor contracting into one single contraction also provides the possibility of efficiently computing on NISQ-era quantum computers.

 \begin{acknowledgments}

    This work was supported by USA Office of Naval Research Global (N62909-19-1-2047)
    and SUTD-ZJU Visiting Professor (VP 201303). T.Z acknowledges the support of
    Singapore Ministry of Education PhD Research Scholarship.

 \end{acknowledgments}

  \appendix
  \section{Mathematical proof for stack operation of two MPS}
  \label{app:mathproof}
  For simplicity, we consider performing the stack operation of two MPS: $\ket{\psi_\alpha}, \ket{\psi_\beta}$, which can be written as
       \begin{align}
           \label{eq:stackoperation}
            & \begin{bmatrix}
               \ket{\psi_\alpha}, & \ket{\psi_\beta}
           \end{bmatrix}=\ket{\psi_\alpha}\otimes\ket{0}+\ket{\psi_\beta}\otimes\ket{1}
       \end{align}
       and
       \begin{align}
            & |\psi_{\alpha}\rangle = \sum_{\{\bm{\sigma}\}}\mathrm{Tr}\left[M_{\alpha}^{\sigma_0}M_{\alpha}^{\sigma_1}\dots M_{\alpha}^{\sigma_{n-1}}\right]|\sigma_0,\dots,\sigma_{n-1}\rangle \\ \nonumber
            & |\psi_{\beta}\rangle = \sum_{\{\bm{\sigma}\}}\Tr\left[M_{\beta}^{\sigma_0}M_{\beta}^{\sigma_1}\dots M_{\beta}^{\sigma_{n-1}}\right]|\sigma_0,\dots,\sigma_{n-1}\rangle
       \end{align}
       are the two MPS's. We seek to find a new MPS $\ket{\phi}$ which exactly represents the stacked operation from Eq.~\eqref{eq:stackoperation}. We denote the new MPS as
       \begin{align}
           \label{eq:stackedMPSphi}
            & |\phi\rangle  = \sum_{\{\bm{\sigma}\}}\Tr\left[\tilde{M}^{\sigma_0}\tilde{M}^{\sigma_1}\dots \tilde{M}^{\sigma_{n-1}}\right]|\sigma_0,\dots,\sigma_{n-1}\rangle
       \end{align}
       To see the detailed structure of each $\tilde{M}^{\sigma_i}$ in the above expression, we expand both terms on the right-hand side of Eq.~\eqref{eq:stackoperation}
       \begin{align}
           \label{eq:lhs1}
            & |\psi_\alpha\rangle|0\rangle\\ \nonumber
            & =\sum_{\sigma_0,\dots,\sigma_{n-1}}\Tr\left[M_{\alpha}^{\sigma_0}M_{\alpha}^{\sigma_1}\dots M_{\alpha}^{\sigma_{n-1}}\right]|\sigma_0,\dots,\sigma_{n-1}\rangle|0\rangle                             \\ \nonumber
            & =\sum_{\sigma_0,\dots,{\sigma_{n}}}\Tr\left[M_{\alpha}^{\sigma_0}M_{\alpha}^{\sigma_1}\dots M_{\alpha}^{\sigma_{n-1}}\right]\delta_0^{\sigma_{n}}|\sigma_0,\dots,\sigma_{n-1}\rangle|\sigma_n\rangle \\ \nonumber
            &  =\sum_{\sigma_0,\dots,{\sigma_{n}}}\Tr\left[M_{\alpha}^{\sigma_0}M_{\alpha}^{\sigma_1}\dots M_{\alpha}^{\sigma_{n-1}}\Delta_0^{\sigma_{n}}\right]|\sigma_0,\dots,\sigma_{n-1},\sigma_n\rangle
       \end{align}
       where $\Delta_0^{\sigma_n}=\delta_0^{\sigma_n}\mathbb{I}$, and $\mathbb{I}$ is the identity matrix of which the row size is the same as the column size of the previous site matrix $M_\sigma^{\sigma_{n-1}}$.

       Similarily, we have
       \begin{align}
           \label{eq:lhs2}
            & |\psi_\beta\rangle|1\rangle\\ \nonumber
            &=\sum_{\sigma_0,\dots,{\sigma_{n}}}\Tr\left[M_{\beta}^{\sigma_0}M_{\beta}^{\sigma_1}\dots M_{\beta}^{\sigma_{n-1}}\Delta_1^{\sigma_{n}}\right]|\sigma_0,\dots,\sigma_{n-1},\sigma_n\rangle
       \end{align}
       By summing up Eq.~\eqref{eq:lhs1} and \eqref{eq:lhs2}, we obtain
       \begin{align}
            & \ket{\psi_\alpha}\otimes\ket{0}+\ket{\psi_\beta}\otimes\ket{1}                                    \\ \nonumber
            & =\sum_{\{\bm{\sigma}\}}\Tr\left[M^{\sigma_0}_{\alpha}M^{\sigma_1}_{\alpha}\dots M^{\sigma_{n-1}}_{\alpha}\Delta_0^{\sigma_n}\right]|\sigma_0,\dots,\sigma_{n-1},\sigma_{n}\rangle    \\ \nonumber
            & +\sum_{\{\bm{\sigma}\}}\Tr\left[M^{\sigma_0}_{\beta}M^{\sigma_1}_{\beta}\dots M^{\sigma_{n-1}}_{\beta}\Delta_1^{\sigma_n}\right]|\sigma_0,\dots,\sigma_{n-1},\sigma_{n}\rangle       \\ \nonumber
            & =\sum_{\{\bm{\sigma}\}}\Tr\left[ M^{\sigma_0}_{\alpha}M^{\sigma_1}_{\alpha}\dots M^{\sigma_{n-1}}_{\alpha}\Delta_0^{\sigma_n} \right.                                                         \\ \nonumber &\left.+M^{\sigma_0}_{\beta}M^{\sigma_1}_{\beta}\dots M^{\sigma_{n-1}}_{\beta}\Delta_1^{\sigma_n}\right]\ket{\sigma_0,\sigma_1,\dots,\sigma_{n-1},\sigma_{n}} \\ \nonumber
            & =\sum_{\{\bm{\sigma}\}}\Tr\left\{
           \left[\begin{array}{c}
                   M_{\alpha}^{\sigma_{0}} \\ M_{\beta}^{\sigma_{0}}
               \end{array}\right]^T
           \left[\begin{array}{cc}
                   M_{\alpha}^{\sigma_{0}} & 0                      \\
                   0                       & M_{\beta}^{\sigma_{0}} \\
               \end{array}\right]
           \cdots \right.                                                                                            \\ \nonumber
            & \left.\left[\begin{array}{cc}
                   M_{\alpha}^{\sigma_{n-1}} & 0                        \\
                   0                         & M_{\beta}^{\sigma_{n-1}} \\
               \end{array}\right]
           \left[\begin{array}{cc}
                   \Delta_0^{\sigma_n} & 0                   \\
                   0                   & \Delta_1^{\sigma_n} \\
               \end{array}\right]\right\}
           \ket{\sigma_0,\sigma_1,\dots,\sigma_{n-1},\sigma_{n}}\\ \nonumber
            & =\sum_{\{\bm{\sigma}\}}\Tr\left[\tilde{M}^{\sigma_0}\tilde{M}^{\sigma_1}\dots \tilde{M}^{\sigma_{n-1}}\tilde{\Delta}^{\sigma_{n}}\right]\ket{\sigma_0,\sigma_1,\dots,\sigma_{n-1},\sigma_{n}} \\ \nonumber
            & =\sum_{\{\bm{\sigma}\}}\Tr\left[\tilde{M}^{\sigma_0}\tilde{M}^{\sigma_1}\dots \tilde{M}^{\sigma_{n-1}}\right]\ket{\sigma_0,\sigma_1,\dots,\sigma_{n-1},\sigma_{n}}
       \end{align}
       Note that for the first site $\sigma_0$, the resulting matrix $\tilde{M}^{\sigma_0}$ is stacked in row geometry [see the first site of the resulting stacked MPS in Fig.~\ref{fig:stackoperation_result}], all the other matrices $\tilde{M}^{\sigma_j} (j=1,\cdots,n-1)$ are block diagonal [Fig.~\ref{fig:stackoperation_result}], as the stack operation is essentially equivalent to a direct sum of local matrices.
       In the final step of the above equation, to maintain the form of the stacked MPS as shown in Eq.~\eqref{eq:stackedMPSphi}, we have absorbed $\tilde{\Delta}^{\sigma_{n}}$ into $\tilde{M}^{\sigma_{n-1}}$ and reformulate it as
       \begin{align}
            & \tilde{M}^{\sigma_{n-1}} \rightarrow \tilde{M}^{\sigma_{n-1}}\tilde{\Delta}^{\sigma_{n}} \\ \nonumber
            & = \left[\begin{array}{cc}
                   M_{\alpha}^{\sigma_{n-1}} & 0                        \\
                   0                         & M_{\beta}^{\sigma_{n-1}} \\
               \end{array}\right]
           \left[\begin{array}{cc}
                   \Delta_0^{\sigma_n} & 0                   \\
                   0                   & \Delta_1^{\sigma_n} \\
               \end{array}\right]                                                     \\ \nonumber
            & =\left[\begin{array}{cc}
                   M_{\alpha}^{\sigma_{n-1}}\Delta_0^{\sigma_n} & 0                                           \\
                                                                & M_{\beta}^{\sigma_{n-1}}\Delta_1^{\sigma_n} \\
               \end{array} \right]
       \end{align}
       which is still block diagonal, but each block with size $D\times 1$.

       The above process is the stack operation for two MPS, and it is easy to extrapolate to the stack operation of $n$ MPS. For simplicity, we assign MPS with the same auxiliary dimension $D$ for each bond. It is not necessary as they can be different. One only needs to change the diagonal block indices from $(D,D)$ to $(D_i,D_j)$.

\providecommand{\noopsort}[1]{}\providecommand{\singleletter}[1]{#1}%

\end{document}